\documentclass[journal]{IEEEtran}

\usepackage{bbding}

\usepackage{cite}

\usepackage{multirow}

\ifCLASSINFOpdf
   \usepackage[pdftex]{graphicx}
  \graphicspath{{../pdf/}{../jpeg/}}
  \DeclareGraphicsExtensions{.pdf,.jpeg,.png}
\else
  \usepackage[dvips]{graphicx}
  \graphicspath{{../eps/}}
  \DeclareGraphicsExtensions{.eps}
\fi
\usepackage{amsmath}
\usepackage{algorithmic}

\usepackage{array}

\ifCLASSOPTIONcompsoc
  \usepackage[caption=false,font=normalsize,labelfont=sf,textfont=sf]{subfig}
\else
  \usepackage[caption=false,font=footnotesize]{subfig}
\fi
\usepackage{fixltx2e}
\usepackage{url}

\begin{document}

%
\title{Road Extraction by Deep Residual U-Net}
%
%
%

\author{Zhengxin Zhang$^{\dagger}$, 
	Qingjie Liu$^{\dagger *}$,~\IEEEmembership{Member,~IEEE}
	and Yunhong Wang,~\IEEEmembership{Senior Member,~IEEE}
\thanks{This work was supported by the Natural Science Foundation of China (NSFC) under Grant 61601011.}
\thanks{Authors are with the State Key Laboratory of Virtual Reality Technology and Systems under the School of Computer Science and Engineering, Beihang University, Beijing 100191, China.}
\thanks{$^{\dagger}$ Authors contribute equally to this letter. }
\thanks{$^{*}$ Corresponding author: Qingjie Liu (qingjie.liu@buaa.edu.cn).}
\thanks{Manuscript received **, 2017; revised **, 2017.}
}

%
%

\markboth{IEEE GEOSCIENCE AND REMOTE SENSING LETTERS}%
{Shell \MakeLowercase{\textit{et al.}}: Bare Demo of IEEEtran.cls for IEEE Journals}
%



\maketitle

\begin{abstract}
Road extraction from aerial images has been a hot research topic in the field of remote sensing image analysis. In this letter, a semantic segmentation neural network which combines the strengths of residual learning and U-Net is proposed for road area extraction. The network is built with residual units and has similar architecture to that of U-Net. The benefits of this model is two-fold: first, residual units ease  training of deep networks. Second, the rich skip connections within the network could facilitate information propagation, allowing us to design networks with fewer parameters however better performance. We test our network on a public road dataset and compare it with U-Net and other two state of the art deep learning based road extraction methods. The proposed approach outperforms all the comparing methods, which demonstrates its superiority over recently developed state of the arts. 
\end{abstract}


%

\section{Introduction}
\IEEEPARstart{R}{oad} extraction is one of the fundamental tasks in the field of remote sensing. It has a wide range of applications such as automatic road navigation, unmanned vehicles, urban planning, and geographic information update, etc. Although it has been received considerable attentions in the last decade, road extraction from high resolution remote sensing images is still a challenging task because of the noise, occlusions and complexity of the background in raw remote sensing imagery.

A variety of methods have been proposed to extract roads from remote sensing images in recent years. Most of these methods can be divided into two categories: road area extraction and road centerline extraction. Road area  extraction~\cite{Xin2009Road,mnih2010learning,Unsalan2012Road,Cheng2015Urban,Saito2016Multiple,Alshehhi2017Hierarchical} can generate pixel-level labeling of roads, while road centerline extraction~\cite{Liu2015Main,Sujatha2015Connected} aims at detecting skeletons of a road. There are also methods extract both road areas and centerline, simultaneously~\cite{Cheng2017Automatic}. Since road centerline can be easily obtained from road areas using algorithms such as morphological thinning~\cite{Cheng2016Road}, this letter focuses on road area extraction from high resolution remote sensing images.

Road area extraction can be considered as a segmentation or pixel-level classification problem. For instance, Song and Civco~\cite{Song2004Road} proposed a method utilizing shape index feature and support vector machine (SVM) to detect road areas. Das et al.~\cite{Das2011Use} exploited two salient features of roads and designed a multistage framework to extract roads from high resolution multi-spectral images using probabilistic SVM. Alshehhi and Marpu~\cite{Alshehhi2017Hierarchical} proposed an unsupervised road extraction method based on hierarchical graph-based image segmentation.

Recent years have witnessed great progress in deep learning. Methods based on deep neural networks have achieved state-of-the-art performance on a variety of computer vision tasks, such as scene recognition~\cite{Zhou2014Learning} and object detection~\cite{Ren2017Faster}. Researchers in remote sensing community also seek to leverage the power of deep neural networks to solve the problems of interpretation and understanding of remote sensing data~\cite{mnih2010learning,mnih2012learning,zhang2016cnn,Saito2016Multiple,Zhang2016Deep,zhang2016functional}. These methods provide better results than traditional ones, showing great potential of applying deep learning techniques to analyze remote sensing tasks.

In the field of road extraction, one of the first attempts of applying deep learning techniques was made by Mnih and Hinton~\cite{mnih2010learning}. They proposed a method employing restricted Boltzmann machines (RBMs) to detect road areas from high resolution aerial images. To achieve better results, a pre-processing step before the detection and a post-processing step after the detection were applied. The pre-processing was deployed to reduce the dimensionality of the input data. The post-processing was employed to remove disconnected blotches and fill in the holes in the roads. Different from Mnih and Hinton's method~\cite{mnih2010learning} that use RBMs as basic blocks to built deep neural networks, Saito et al.~\cite{Saito2016Multiple} employed Convolutional Neural Network (CNNs) to extract buildings and roads directly from raw remote sensing imagery. This method achieves better results than Mnih and Hinton's method~\cite{mnih2010learning} on the Massachusetts roads dataset.

Recently, lots of works have suggested that a deeper network would have better performance~\cite{szegedy2015going,simonyan2014very}. However, it is very difficult to train a very deep architecture due to problems such as vanishing gradients. To overcome this problem, He et al.~\cite{resnet2015deep} proposed the deep residual learning framework that utilize an identity mapping~\cite{resnet2016} to facilitate  training. Instead of using skip connection in Fully Convolutional Networks (FCNs)~\cite{FCN2015fully}, Ronneberger et al.~\cite{U-NET2015} proposed the U-Net that concatenate  feature maps from different levels to improve segmentation accuracy. U-Net combines low level detail information and high level semantic information, thus achieves promising performance on biomedical image segmentation~\cite{U-NET2015}. 

Inspired by the deep residual learning~\cite{resnet2015deep} and U-Net~\cite{U-NET2015}, in this letter we propose the deep residual U-Net, an architecture that take advantage of strengths from both deep residual learning and U-Net architecture. The proposed deep residual U-Net (ResUnet) is built based on the architecture of U-Net. The differences between our deep ResUnet and U-Net are in two-fold. First, we use residual units instead of plain neural units as basic blocks to build the deep ResUnet. Second, the cropping operation is unnecessary thus removed from our network, leading to a much more elegant architecture and better performance. 


\vspace{-0.3cm}
\section{Methodology}
\label{sec:methodology}

\subsection{Deep ResUnet}
\subsubsection{U-Net}
In semantic segmentation, to get a finer result, it is very important to use low level details while retaining high level semantic information~\cite{FCN2015fully,U-NET2015}. However, training such a deep neural network is very hard especially when only limited training samples are available. One way to solve this problem is employing a pre-trained network then fine-tuning it on the target dataset, as done in \cite{FCN2015fully}. Another way is employing extensive data augmentation, as done in U-Net~\cite{U-NET2015}. In addition to data augmentation, we believe the architecture of  U-Net also contributes to relieving the training problem. The intuition behind this is that copying low level features to the corresponding high levels actually creates a path for information propagation allowing signals propagate between low and high levels in a much easier way, which not only facilitating backward propagation during training, but also compensating low level finer details to high level semantic features. This somehow shares similar idea to that of residual neural network~\cite{resnet2015deep}. In this letter, we show that the performance of U-Net can be further improved by substituting the plain unit with a residual unit. 
\begin{figure}[t!]
	\begin{center}
		\includegraphics[width=0.87\columnwidth]{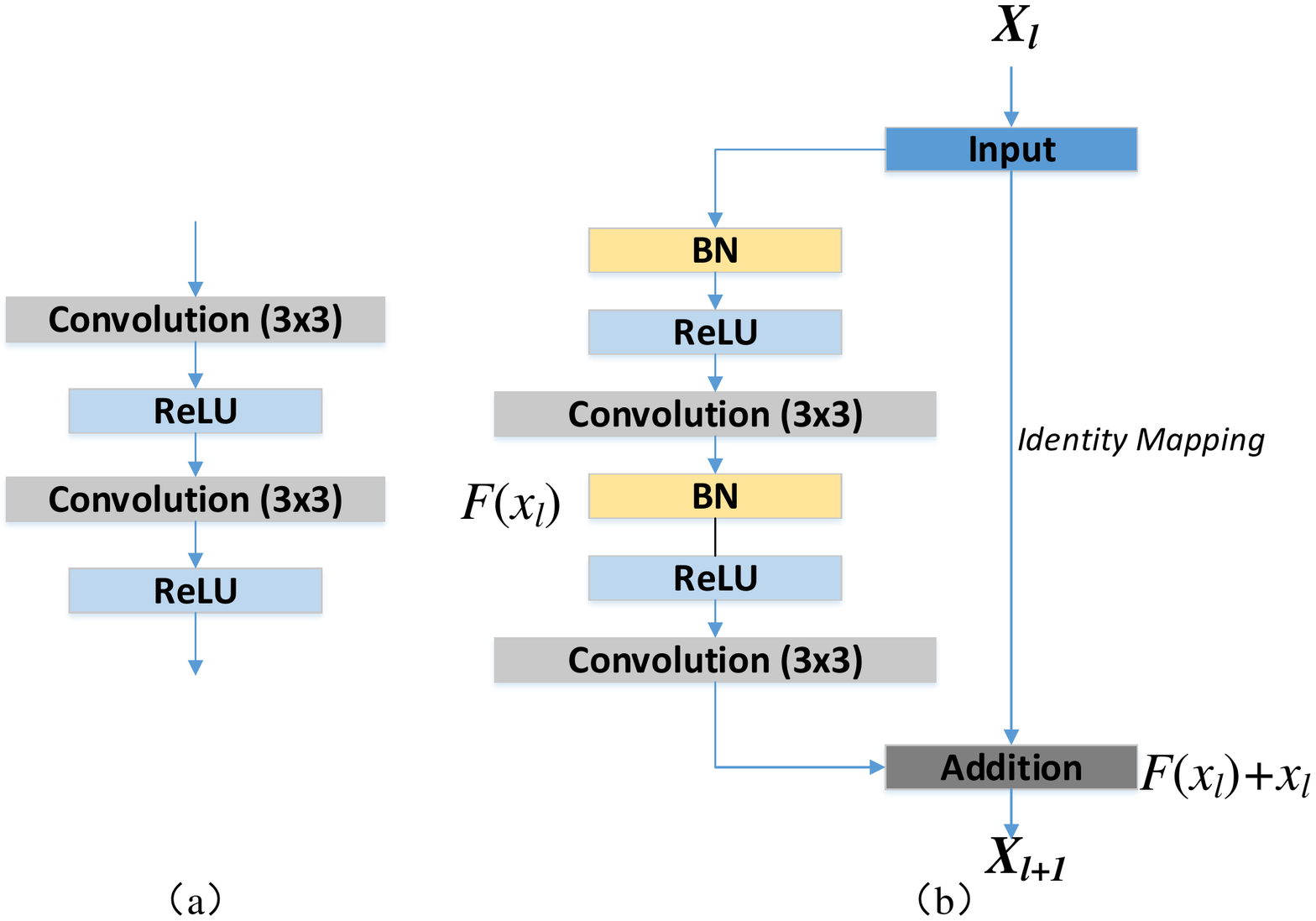}
		\caption{Building blocks of neural networks. (a) Plain neural unit used in U-Net and (b) residual unit with identity mapping used in the proposed ResUnet.}
		\label{Fig:ResNet Block}
	\end{center}
	\vspace{-0.6cm}
\end{figure}
\subsubsection{Residual unit}
Going deeper would improve the performance of a multi-layer neural network, however could hamper the training, and a degradation problem maybe occur~\cite{resnet2015deep}. To overcome these problems, He et al.~\cite{resnet2015deep} proposed the residual neural network to facilitate  training and address the degradation problem. The residual neural network consists of a series of stacked residual units. Each residual unit can be illustrated as a general form:
\begin{equation}\label{Equ:Residual Uint}
\begin{split}
\mathbf{y}_{l}\ \ \ & = h(\mathbf{x}_{l})+\mathcal{F}(\mathbf{x}_{l}, \mathcal{W}_{l}), \\
\mathbf{x}_{l+1} & = f(\mathbf{y}_{l}),
\end{split}
\end{equation}
where $\mathbf{x}_{l}$ and $\mathbf{x}_{l+1}$ are the input and output of the $l$-th residual unit, $\mathcal{F}(\cdot)$ is the residual function, $f(\mathbf{y}_l)$ is activation function and $h(\mathbf{x}_{l})$ is a identity mapping function, a typical one is  $h(\mathbf{x}_{l}) = \mathbf{x}_{l}$. Fig.~\ref{Fig:ResNet Block} shows the difference between a plain and residual unit. There are multiple combinations of batch normalization (BN), ReLU activation and convolutional layers in a residual unit. He et al. presented a detailed discussion on impacts of different combinations in \cite{resnet2016} and suggested a full pre-activation design as shown in  Fig.~\ref{Fig:ResNet Block}(b). In this work, we also employ full pre-activation residual unit to build our deep residual U-Net. 
\begin{figure}[tbp!]
	\begin{center}
		\includegraphics[width=0.87\columnwidth]{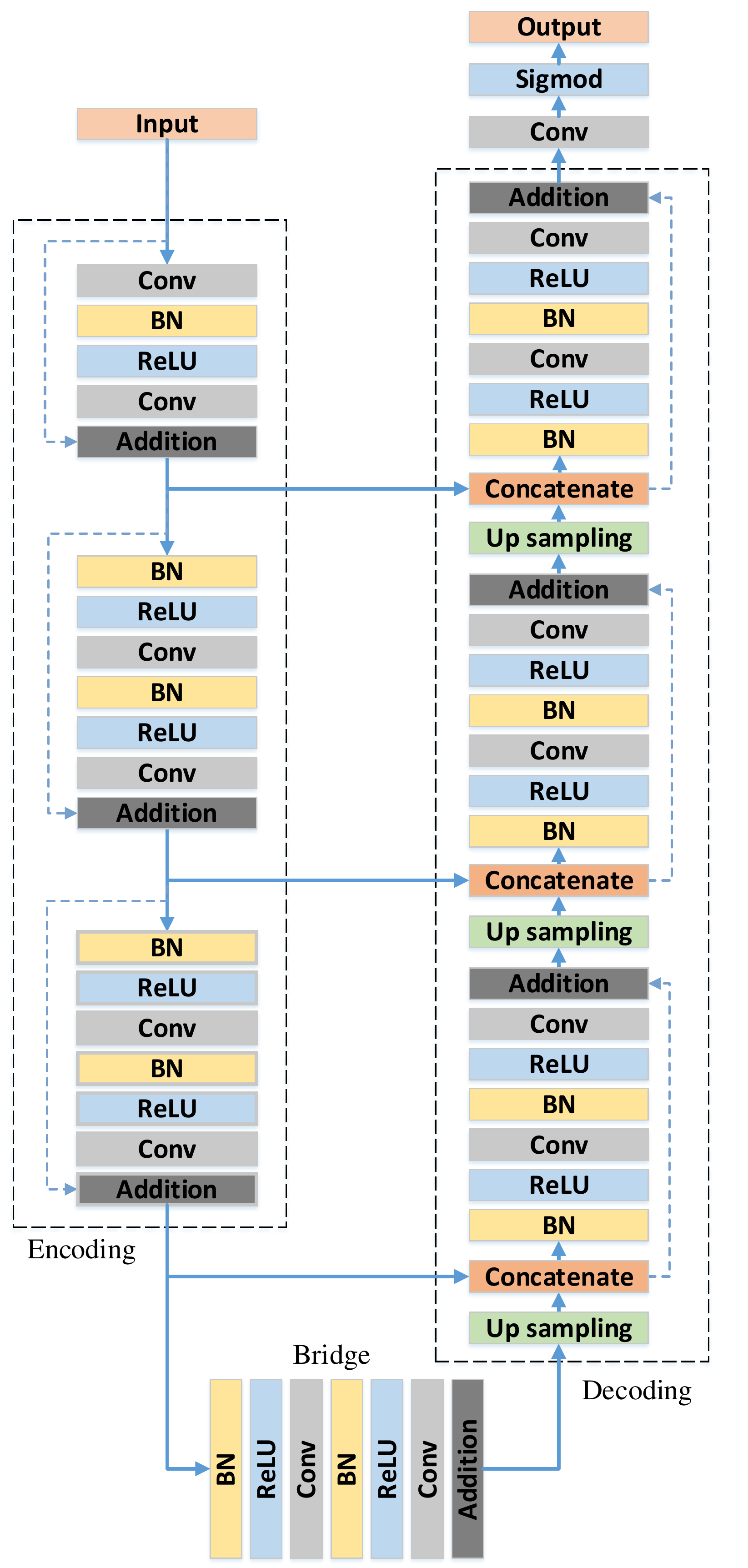}
		\caption{The architecture of the proposed deep ResUnet.}
		\label{Fig:Deep Res-U-Net}
	\end{center}
	\vspace{-0.7cm}
\end{figure}

\subsubsection{Deep ResUnet}
Here we propose the deep ResUnet, a semantic segmentation neural network which combines strengths of both U-Net and residual neural network. This combination bring us two benefits: 1) the residual unit will ease training of the network; 2) the skip connections within a residual unit and between low levels and high levels of the network will facilitate information propagation without degradation, making it possible to design a neural network with much fewer parameters however could achieve comparable ever better performance on semantic segmentation. 

In this work, we utilize a 7-level architecture of deep ResUnet for road area extraction, as shown in Fig.~\ref{Fig:Deep Res-U-Net}. The network comprises of three parts: encoding, bridge and decoding.\footnote{U-Net used  ``contracting" and ``expansive" paths to denote the feature extraction and up-convolution stages of the network. In this letter, we prefer the terms encoding and decoding because we think it is more meaningful and easer to understand.} The first part encodes the input image into compact representations. The last part recovers the representations to a pixel-wise categorization, i.e. semantic segmentation. The middle part serves like a bridge connecting the encoding and decoding paths. All of the three parts are built with residual units which consist of two $3\times3$ convolution blocks and an identity mapping. Each convolution block includes a BN layer, a ReLU activation layer and a convolutional layer. The identity mapping connects input and output of the unit. 

Encoding path has three residual units. In each unit, instead of using pooling operation to downsample the feature map size, a stride of 2 is applied to the first convolution block to reduce the feature map by half. Correspondingly, decoding path composes of three residual units, too. Before each unit, there is an up-sampling of feature maps from lower level and a concatenation with the feature maps from the corresponding encoding path. After the last level of decoding path, a $1\times1$ convolution and a sigmod activation layer is used to project the multi-channel feature maps into the desired segmentation. In total we have 15 convolutional layers comparing with 23 layers of U-Net. It is worth noting that the indispensable cropping in U-Net is unnecessary in our network. The parameters and output size of each step are presented in Table~\ref{Table:Feature Size}. 

\begin{table}[!htb]
	\tiny
	\centering
	\caption{The network structure of ResUnet.}	
	\label{Table:Feature Size}	
	\begin{tabular}{ccllcl}
		\hline
		\hline
		& Unit level  & Conv layer & Filter  & Stride &  Output size\\
		\hline
		Input                      &        &     &   &    &  $224\times 224 \times 3$ \\
		\hline	
		\multirow{6}{*}{Encoding}
		& \multirow{2}{*}{Level 1}& Conv 1  & $3\times 3/64$ & 1 &  $224\times 224 \times 64$ \\
		&        & Conv 2  &  $3\times 3/64$ & 1 &  $224\times 224 \times 64$ \\
		\cline{2-6}			
		& \multirow{2}{*}{Level 2}& Conv 3  &  $3\times 3/128$ & 2 &  $112\times 112 \times 128$ \\
		&        & Conv 4  &  $3\times 3/128$ & 1 &  $112\times 112 \times 128$ \\
		\cline{2-6}
		& \multirow{2}{*}{Level 3}& Conv 5  &  $3\times 3/256$ & 2 &  $56\times 56 \times 256$ \\
		&        & Conv 6  &  $3\times 3/256$ & 1 &  $56\times 56 \times 256$ \\
		\cline{2-6}						
		\hline
		\multirow{2}{*}{Bridge}
		&\multirow{2}{*}{Level 4}         &Conv 7  &  $3\times 3/512$ & 2 &  $28\times 28 \times 512$ \\
		&	      &Conv 8 &  $3\times 3/512$ & 1 &  $28\times 28 \times 512$\\
		\hline
		\multirow{6}{*}{Decoding}	
		& \multirow{2}{*}{Level 5} &Conv 9 &  $3\times 3/256$ & 1 &  $56\times 56 \times 256$ \\
		&        &Conv 10 &  $3\times 3/256$ & 1 & $56\times 56 \times 256$ \\
		\cline{2-6}
		& \multirow{2}{*}{Level 6} &Conv 11 &  $3\times 3/128$ & 1 &  $112\times 112 \times 128$ \\
		&        &Conv 12 &  $3\times 3/128$ & 1 &  $112\times 112 \times 128$ \\
		\cline{2-6}			
		& \multirow{2}{*}{Level 7} &Conv 13 &  $3\times 3/64$ & 1 &   $224\times 224 \times 64$ \\
		&        &Conv 14 &  $3\times 3/64$ & 1 &  $224\times 224 \times 64$ \\
		\hline
		Output                      &        &Conv 15 &  $1\times 1$ & 1 &  $224\times 224 \times 1$ \\		
		\hline
		\hline
	\end{tabular}
	\vspace{-0.6cm}
\end{table}

\subsection{Loss function}
Given a set of training images and the corresponding ground truth segmentations $\{I_i,s_i\}$, our goal is to estimate parameters $W$ of the network, such that it produce accurate and robust road areas. This is achieved through minimizing the loss between the  segmentations generated by $Net(I_i;W)$ and the ground truth $s_i$. In this work, we use Mean Squared Error (MSE) as the loss function:
\begin{equation}\label{Equ:mse}
\mathcal{L}(W) = \frac{1}{N}\sum\limits^{N}_{i=1}||Net(I_i;W) - s_i||^2,
\end{equation}
where $N$ is the number of the training samples. We use the stochastic gradient descent (SGD) to train our network. One should know that other loss functions that are derivable can also be used to train the network. For instance, U-Net adopted pixel-wise cross entropy as loss function to optimize the model. 

\subsection{Result refinement}

The input and output of our semantic segmentation network have the same size in width and height, both are $224\times224$. The pixels near  boundaries of the output have lower accuracy than center ones due to zero padding in the convolutional layer. To get a better result, we use an overlap strategy to produce the segmentation results of a large image. The input sub-images are cropped from the original image with an overlap of $o$ ($o=14$ in our experiments). The final results are obtained by stitching all sub-segmentations together. The values in the overlap regions are averaged. 

\vspace{-0.2cm}
\section{Experiments}
\label{sec:experiment}

To demonstrate the accuracy and efficiency of the proposed deep ResUnet, we test it on Massachusetts roads dataset\footnote{https://www.cs.toronto.edu/\~{}vmnih/data/} and compare it with three state of the art methods, including Mnih's~\cite{mnih2010learning} method, Saito's method~\cite{Saito2016Multiple} and U-Net~\cite{U-NET2015}. 

\vspace{-0.2cm}
\subsection{Dataset}
The Massachusetts roads dataset was built by Mihn et al.~\cite{mnih2010learning}. The dataset consists of 1171 images in total, including 1108 images for training, 14 images for validation and 49 images for testing. The size of all the images in this dataset is $1500\times1500$ pixels with a resolution of 1.2 meter per pixel. This dataset roughly covers 500 km$^2$ space crossing from urban, sub-urban to rural areas and a wide range of ground objects including roads, rivers, sea, various buildings, vegetations, schools, bridges, ports, vehicles, etc. In this work, we train our network on the training set of this dataset and report results on its test set.

\subsection{Implementation details}
The proposed model was implemented using Keras~\cite{chollet2015keras} framework and optimized by minimizing Eqn.~\ref{Equ:mse} through SGD algorithm. There are 1108 training images sized $1500\times1500$ available for training. Theoretically, our network can take arbitrary size image as input, however it will need amount of GPU memory to store the feature maps. In this letter, we utilize fixed-sized training images ($224\times224$ as described in Table~\ref{Table:Feature Size}) to train the model. These training images are randomly sampled from the original images. At last, 30,000 samples are generated and fed into the network to learn the parameters. It should be noted that, no data augmentation is used during training. We start training the model with a mini-batch size of 8 on a NVIDIA Titan 1080 GPU. The learning rate was initially set to 0.001 and reduced by a factor of 0.1 in every 20 epochs. The network will converge in 50 epochs. 
\subsection{Evaluation metrics}
The most common metrics for evaluating a binary classification method are precision and recall. In remote sensing, these metrics are also called correctness and completeness. The precision is the fraction of predicted road pixels which are labeled as roads and the recall is the fraction of all the labeled road pixels that are correctly predicted.

Because of the difficulty in correctly labeling all the road pixels, Mnih et al.~\cite{mnih2010learning} introduced the relaxed precision and recall scores~\cite{Ehrig2005Relaxed} into road extraction. The relaxed precision is defined as the fraction of number of pixels predicted as road within a range of $\rho$ pixels from pixels labeled as road. The relaxed recall is the fraction of number of pixels labeled as road that are within a range of $\rho$ pixels from pixels predicted as road. In this experiment, the slack parameter $\rho$ is set to 3, which is consistent with previous studies\cite{mnih2010learning,Saito2016Multiple}. We also report break-even points of different methods. The break-even point is defined as the point on the relaxed precision-recall curve where its precision value equals its recall value. In other words, break-even point is the intersection of precision-recall curve and line $y=x$. 

\subsection{Comparisons}
\begin{figure}[!t]
\begin{center}
		\includegraphics[width=1\columnwidth]{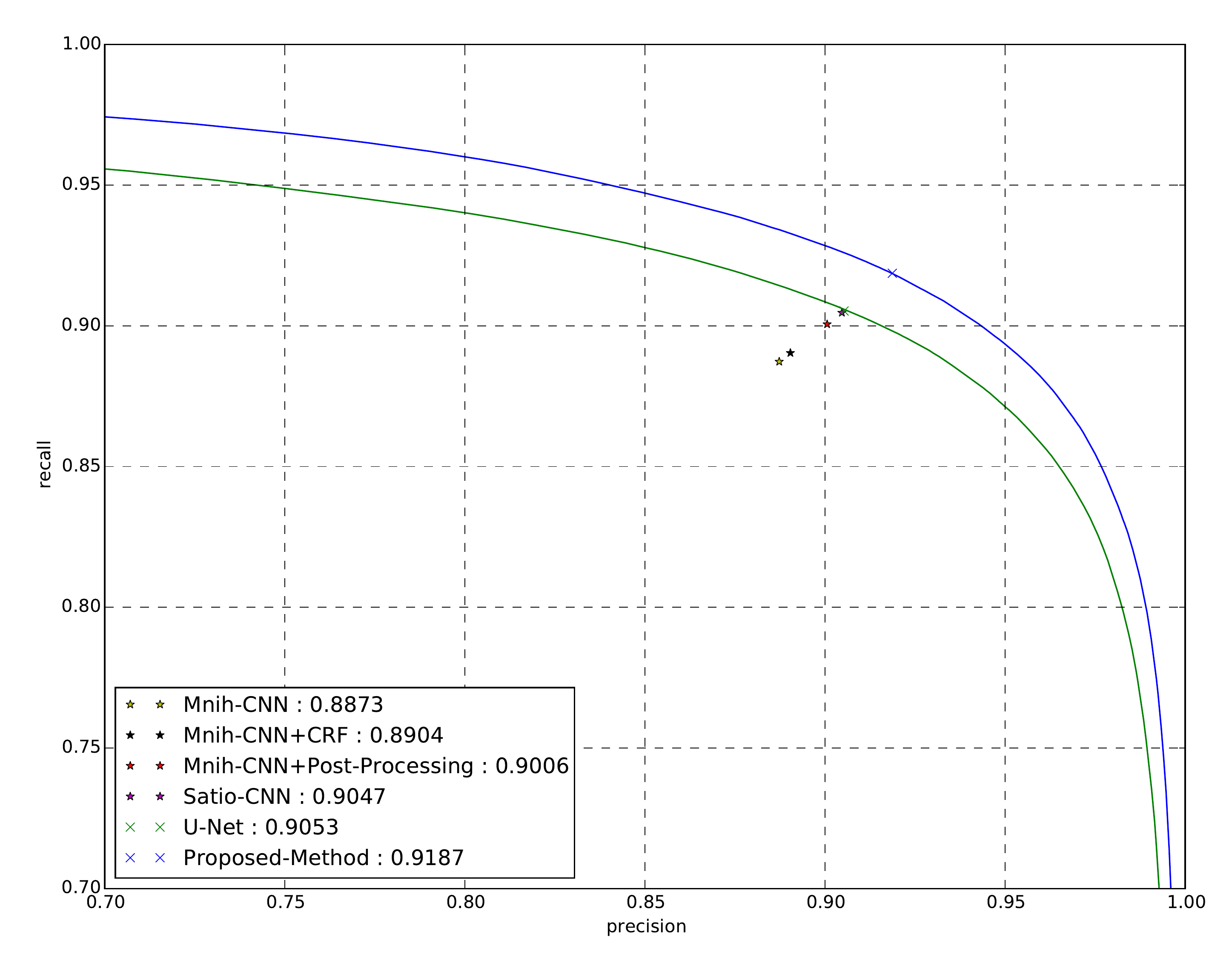}
	    \caption{The relaxed precision-recall curves of U-Net and the proposed method on Massachusetts roads dataset. The marks `$\star$' and `$\times$' are break-even points of different methods.}
	    \label{Fig:PR}
\end{center}
\vspace{-0.5cm}
\end{figure}

Comparisons with three state of the art deep learning based road extraction methods are conducted on the test set of Massachusetts roads dataset. The break-even points of the proposed and comparing methods are reported in Table~\ref{Table:Value at Breakeven Point}. Fig.~\ref{Fig:PR} presents the relaxed precision-recall curves of U-Net and our network and their break-even points, along with break-even points of comparing methods. It can be seen that our method performs better than all other three approaches in terms of relaxed precision and recall. Although the parameters of our network is only $1/4$ of U-Net (7.8M versus 30.6M), promising improvement are achieved on the road extraction task. 

\begin{table}[!hbp]
\vspace{-0.2cm}
\begin{center}
\caption{Comparisons of the proposed and other three deep learning based road extraction method on Massachusetts roads dataset in terms of breakeven point. A higher breakeven point indicates a better performance in precision and recall. }
\label{Table:Value at Breakeven Point}
\begin{tabular}{l|c}
\hline
Model & Breakeven point\\
\hline
Mnih-CNN\cite{mnih2010learning} & 0.8873  \\
\hline
Mnih-CNN+CRF\cite{mnih2010learning} & 0.8904  \\
\hline
Mnih-CNN+Post-Processing\cite{mnih2010learning} & 0.9006  \\
\hline
Saito-CNN\cite{Saito2016Multiple} & 0.9047 \\
\hline
U-Net\cite{U-NET2015} & 0.9053 \\
\hline
ResUnet & \textbf{0.9187} \\
\hline
\end{tabular}
\end{center}
\vspace{-0.3cm}
\end{table}

\begin{figure*}[ht]
\begin{center}
		\includegraphics[width=1\textwidth]{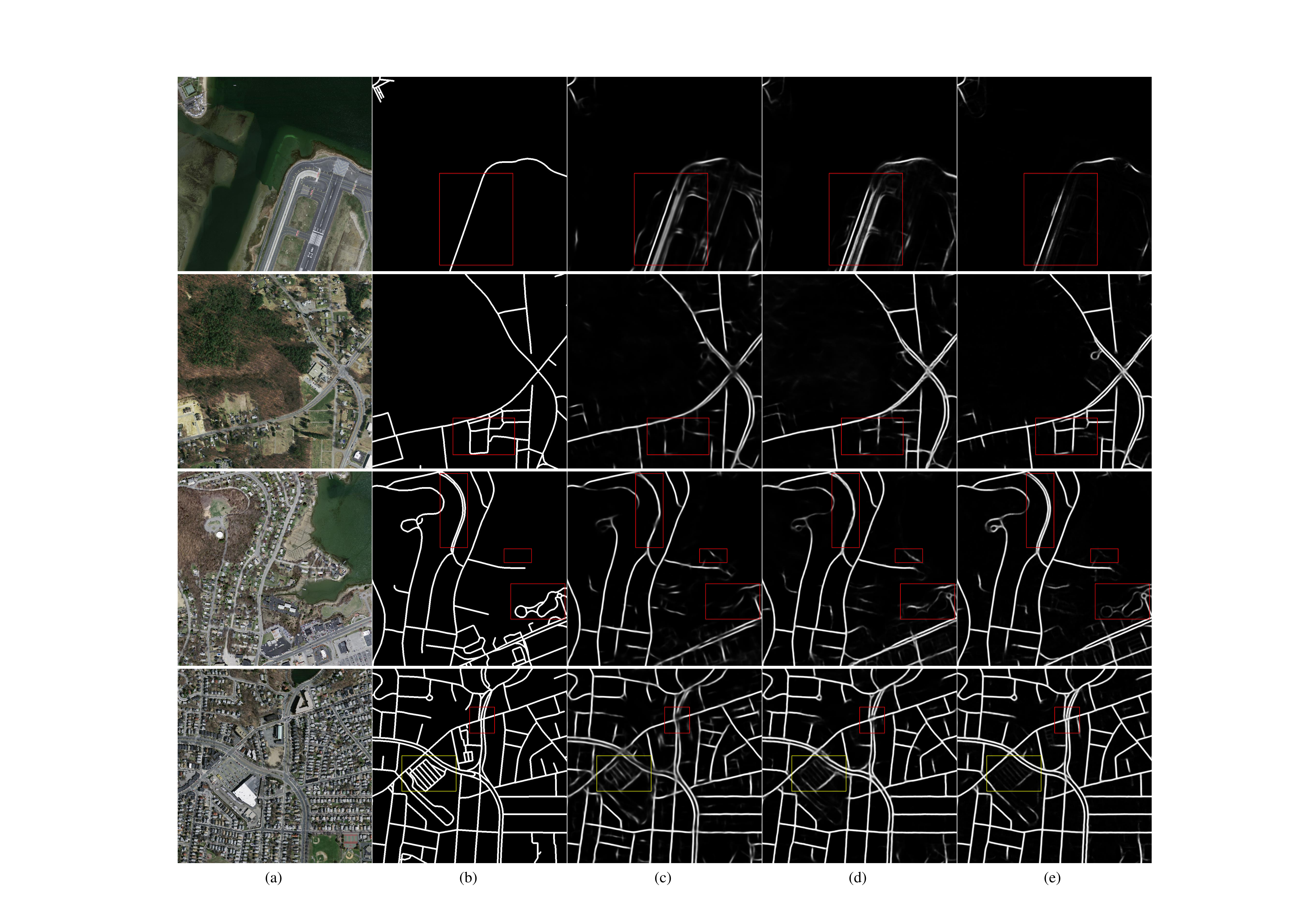}
	    \caption{Example results on the test set of Massachusetts roads dataset. (a) Input image; (b) Ground truth; (c) Saito et al.~\cite{Saito2016Multiple}; (d) U-Net~\cite{U-NET2015}; (e) The proposed ResUnet. Zoom in to see more details.}
	    \label{Fig:Result Comparison_0}
\end{center}
\vspace{-0.6cm}
\end{figure*}

Fig.~\ref{Fig:Result Comparison_0} illustrates four example results of Saito et al.~\cite{Saito2016Multiple}, U-Net~\cite{U-NET2015} and the proposed ResUnet. It can be seen, our method shows cleaner results with less noise than the other two methods. Especially when there are two-lane roads, our method can segmentation each lane with high confidence, generating clean and sharp two-lane roads, while other methods may confuse lanes with each other, as demonstrate in the third row of Fig.~\ref{Fig:Result Comparison_0}. Similarly, in the intersection regions, our method also produces better results.

Context information is very important when analyzing objects with complex structures. Our network considers context information of roads, thus can distinguish roads from similar objects such as building roofs, airfield runways. From the first row of Fig.~\ref{Fig:Result Comparison_0} we can see that, even the runway has very similar features to a highway, our method can successfully segmentation side road from the runway. In addition to this, the context information also make it robust to occlusions. For example, parts of the roads on the rectangle of the second row are covered by trees. Saito's method and U-Net cannot detect road under the trees, however our method labeled them successfully. A failure case is shown in the yellow rectangle of the last row. Our method missed the roads in the parking lot. This is mainly because most of roads in parking lots are not labeled. Therefore, although these roads share the same features to the normal ones, considering the context information our network regard them as backgrounds.

\section{Conclusion}
\label{sec:conslusion}
In this letter, we have proposed the ResUnet for road extraction from high resolution remote sensing images. The proposed network combines the strengths of residual learning and U-Net. The skip connections within the residual units and between the encoding and decoding paths of the network will facilitate information propagations both in forward and backward computations. This property not only ease training but also allows us to design simple yet powerful neural networks. The proposed network outperforms U-Net with only 1/4 of its parameters, as well as other two state of the art deep learning based road extraction methods.

\ifCLASSOPTIONcaptionsoff
  \newpage
\fi

\bibliographystyle{IEEEtran}
\bibliography{IEEEabrv,bibi}

\end{document}